# Defect detection and segmentation in X-Ray images of magnesium alloy castings using the *Detectron2* framework

Francisco Javier Yagüe[1], José Francisco Díez-Pastor[1], Pedro Latorre-Carmona[1], and César Ignacio García Osorio[1]

Departamento de Ingeniería Informática, Universidad de Burgos. Avda. Cantabria s/n, 09006, Burgos, Spain.

**Abstract.** New production techniques have emerged that have made it possible to produce metal parts with more complex shapes, making the quality control process more difficult. This implies that the visual and superficial analysis has become even more inefficient. On top of that, it is also not possible to detect internal defects that these parts could have. The use of X-Ray images has made this process much easier, allowing not only to detect superficial defects in a much simpler way, but also to detect welding or casting defects that could represent a serious hazard for the physical integrity of the metal parts. On the other hand, the use of an automatic segmentation approach for detecting defects would help diminish the dependence of defect detection on the subjectivity of the factory operators and their time dependence variability. The aim of this paper is to apply a deep learning system based on *Detectron2*, a state-of-the-art library applied to object detection and segmentation in images, for the identification and segmentation of these defects on X-Ray images obtained mainly from automotive parts.

## 1 Introduction

In manufacturing, quality control ensures that customers receive *free from defects* products while meeting their needs. It is especially important when these products are going to be used in critical systems, in which a failure can cause serious consequences. One example could be the automobile industry. Competitiveness in the manufacturing market is always a priority. A light car, for instance, is more efficient and gives the user better performance than a heavy one. To reduce the weight of the vehicles, high-strength steel, aluminum (Al), and polymers are being used. But to achieve a significant weight reduction, it is necessary to use materials with lower density. Magnesium alloy pieces have a high technological interest nowadays in different industrial sectors, and specially is an attractive material for automotive use, partly due to some of their properties [23], including their: (a) lightness, Mg is 36% lighter per unit volume than aluminum and 78% lighter than iron; (b) optimal electrical and thermal conductivity; (c) high capability to be recycled, and (d) strength, when alloyed, Mg has the highest strength-to-weight ratio of all the structural metals. But despite the advantages, magnesium is not without its drawbacks, magnesium is more expensive and difficult to operate. During the creation of metal parts, small bubbles or pores may appear that are undetectable by the human eye, even when they are directly on the surface. Detecting these small pores or bubbles is important because it can cause the part to break during operation, which can become a critical problem in the case, for example, of auto parts since they are sometimes in continuous fatigue. There are two main groups of strategies for defect inspection: those that imply a partial or total degradation of the sample to be inspected, and those aimed at analyzing its quality in a remote sensing way. The last group is also called non-destructive examination (NDE) techniques, and they are basically based on acquiring two and three-dimensional images of an object, using techniques like optical, X-ray or ultrasound imaging, just to cite two different types of acquisition *modalities* [13, 15, 25]. Nowadays, the inspection of these pieces is made in a visual and subjective way by personnel working in (for) the compa- nies. They examine the X-Ray images and then determine the position of the defects (if any). This kind of analysis has two main drawbacks: (1) the worker fatigue; (b) the time dependence of the subjective criteria used by the workers to analyse the X-Ray images. To make things worse, it might be unavoidable to have different workers with different subjective criteria to determine the quality of a piece being inspected.

## 2   Problem description

In this paper, we analyze the problem of defect detection and segmentation in the context of high variability images of pieces made by magnesium alloys. This variability is in terms of the different types of samples, and between views of the same piece as well, due to the process of positioning the piece in the X-Ray inspection system and the mechanical imprecision of the positioning system.

A large number of methods for automatic detection and segmentation of defects in X-ray images have been developed in the last few years. The problem of automatic evaluation of the quality of manufactured parts can be addressed using two types of algorithms that are similar but have important differences: Object Detection and Image Segmentation. In both tasks, the aim is at finding certain regions of interest (ROIs) on an image.

### 2.1   Object Detection

The main result of an object detection method is in the form of a bounding box. The algorithms in this category will *create* rectangles around each object of interest in the picture. These rectangles (bounding boxes) are defined by a pair of coordinates (the position of the top left corner), and two values (width and height).

## 2.2   Image Segmentation

Instead of searching for boxes containing the objects of interest, these algorithms tag the image pixel by pixel. The objective of the algorithms is to obtain a transformation of the input image where the background pixels are labeled in one way and the objects of interest in another, obtaining in detail the limits of each object.

In summary, in both type of algorithms input data is a matrix, but in the first case the output is a list of bounding boxes and in the second case the output is also a matrix (mask image) with one value per pixel, containing the assigned category.

## 2.3   Defect detection and segmentation methods

The defect detection and segmentation methods can be classified into: a) image processing methods and b) machine learning methods. In turn, the image processing methods can be classified into: a.1) methods based on the subtraction of a reference image, also called *golden image*, this image can be obtained from the inspected image, using different image processing operations, or it can be obtained automatically from a set of images. a.2) methods based on digital image processing methodologies: The defects are segmented via an automatic threshold or using watershed approaches, among others [22, 5, 6].

Image processing-based methods are older and less flexible. Methods based on machine learning are able to generalize and work in a more flexible set of situations. In the methods of this category, the detection and segmentation of defects becomes a classification problem. The methods can be divided into b.1) classifiers trained with handcrafted features [19, 9] and b.2) deep learning methods, where the features are learned using a neural net [21]. In object detection using deep learning, the most popular architectures are YOLO and Faster RCNN. YOLOv5 has been used in the detection of defects in images manufactured in [20], while Faster RCNN was applied to detect defects in X-Ray images of automotive parts in [10]. When it comes to segmentation, the most popular algorithms are Mask RCNN [11] and U-net [26]. Both has been previously used in manufacturing.

Figure 1 shows two X-Ray images of two different types of typical automotive pieces that are being produced. Defects are visible as irregular internal structures with a higher grayscale level value inside them. Their location, shape and size can be completely random.

We must stress that, to the best of our knowledge, only the work in [10] present results and images that are somehow similar to those presented in this paper, but only for detection, not segmentation. In particular, authors in [10] apply a deep learning based method to detect defects in X-Ray images of automotive parts. Images are of the same *nature* as those used here. The main difference, however, lies in the fact that their aim is to detect defects and create the corresponding bounding boxes defining them, whereas in our case, we are aiming to not only detecting but also segmenting them, as will be presented below.

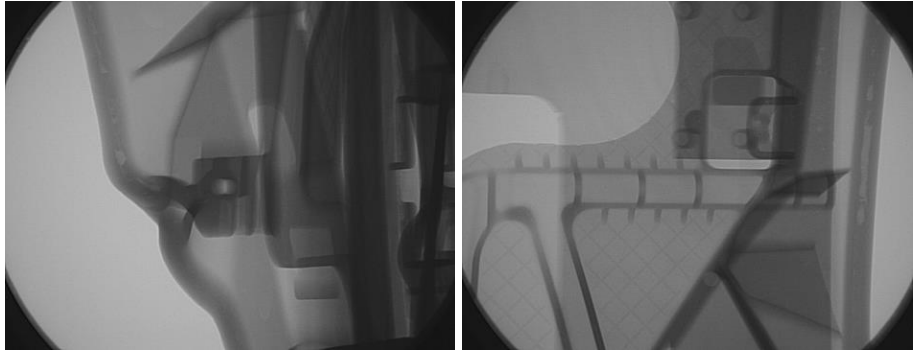

**Fig. 1.** Two X-Ray images showing some defects present during the manufacturing process

## 3   Methodology

Figure 2 shows a flow diagram with the different steps involved in our method. The process can be divided into three main blocks called *image selection and introduction to the platform*, *training stage* and *model application and assessment*.

The first block consists of the image set preparation for its introduction to the platform and previous split up between training and test sets. The second one contains the platform set up and configuration values adjustment before the training phase, and the training itself. Finally, the third block presents the model execution on the test set, metrics assessment and decision whether to finish the process or to modify the training values.

### 3.1   Detectron2

*Detectron2* is a Facebook AI Research (FAIR)'s library [1, 2] for object detection and segmentation in images. It is the successor of *Detectron* and the *MaskRCNN* Benchmark [4], and it is considered to provide state-of-the-art results in both types of tasks. It implements: Mask R-CNN, RetinaNet, Faster R-CNN, Tensor-Mask, DensePose, and other object detection methods. In terms of segmentation, three different types are *supported* : (1) Semantic segmentation, (2) instance segmentation and (3) panoptic segmentation [1]. Panoptic segmentation tries to create a unifying framework between instance and semantic segmentation [14].

### 3.2   Common Objects in Context (COCO) format

In order to use *Detectron2* in a more efficient way, a JSON based image file systems was used, which allows objects of interest to be labelled. This file consists of three parts: (1) *Images*, (2) *Categories* and (3) *Annotations*. All this data was extracted from binary masks or ground truth images that represented the original image on black color and the marked defects on it on white color. The first one of them connects each image file with an identifier and its dimensions [18]. On the other hand, *Categories* allows registering each one of the classes that form part of a group of objects. Finally, *Annotations* contains an object identifier, the category to which it belongs, whether it is a group of objects or not, and also a sequence of the pairs of coordinates of the polygon that delimits the region it occupies, the total area and the bounding box defining coordinates.

---

[1]  In semantic segmentation all ROIs tagged with the same class represent a single entity. In contrast, instance segmentation treats each ROI belonging to the same class as distinct individual instances.

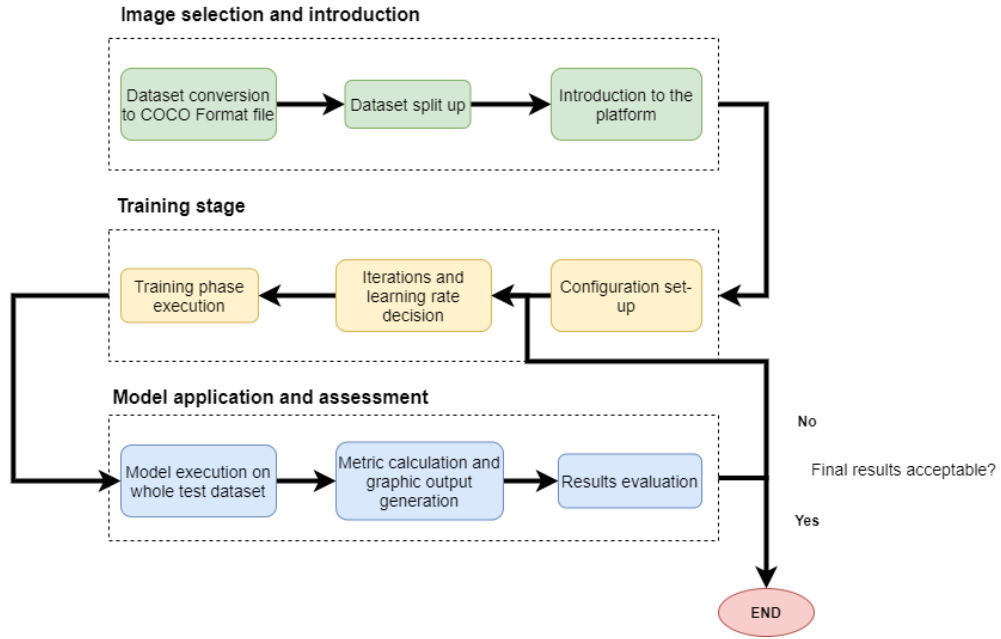

**Fig. 2.** Full process diagram of the defect detection and segmentation framework

### 3.3 Faster R-CNN

Deep learning might be considered as a subgroup of a broader group of machine learning methods that consider an artificial neural network as its core. There is a rich diversity in the type of architectures that would fall under the umbrella of a deep-learning paradigm, including deep neural networks, recurrent neural networks and convolutional neural networks, just to consider a few of them [16]. On the other hand, they have been applied to several fields, including computer vision, natural language processing, and bioinformatics, to cite a few [7,8].

Convolutional Neural Networks (CNNs) [17] are a particular type of deep neural network. Some of their drawbacks are related to the size of the images to be processed and the number of regions that might include the objects of interest. That is the reason why a new method called *Regions with CNN features* (R-CNN) was proposed [12]. It introduces mainly two changes: The extraction of the proposed regions for their posterior combination using a similarity criterion, and the proposal of a series of candidate regions for their final analysis.

Faster R-CNN [24] is an evolved version of R-CNN in terms of speed and simplicity. Other features include: (a) It has a new layer, called *ROI Pooling* and it extracts the feature vectors of an image with the same length; (b) Shared computation: When a *Region of Interest* is processed, it shares the operations using the *ROI pooling* ; (c) Region Proposal Networks [24] are a complete neural network which generates the ROIs at different scales, giving this information to Fast R-CNN; (d) Anchor boxes. They create a reference with the scale and location of possible detections, making it easier to begin the classification process.

### 3.4   Dataset train-test split

A group of 21 images were considered for this study. For all of them, the hand-crafted delineation of all the defects presented in them was made by the authors and verified by personnel from the company. From the complete group of images, 11 were selected for training and 10 for testing. In this process, the user can decide which strategy is the best for each case. However, it is always a good idea to keep a diverse dataset that does not present too similar shapes or defects.

### 3.5   Configuration set up

The training phase is performed by a Python object that takes an initial configuration that can be overwritten by the user to adapt the parameters before launching the process. In particular, its structure is:

– DATASETS.TRAIN: Training dataset in *Common Objects in COntex* (COCO) format that is already converted and introduced to the platform.
– BASE.LR: Learning rate. In all the cases, a value of 0.05 was used
– MAX.ITER: Number of iterations for the training phase, we set this to 600.
– MODEL.ROI.HEADS.NUM CLASSES: Number of different types of classes that contained in the dataset. In our case, we considered just 1.

### 3.6   Training stage

This stage was made using *Google Colaboratory* following the installation, configuration and training instructions that appear in the *Detectron2* official documentation. However, for a more practical approach, an official notebook ready to test the platform is open to the public [3]. Faster R-CNN is composed of two different modules that work together and achieve different tasks.

First, the Region Proposal Network (RPN) proposes regions, and a Fast R-CNN detector that takes input from the RPN and generates both bounding boxes and classes that are detected in the image. RPN is a fully convolutional network aimed to propose a set of rectangular regions that get an associated *objectness* value that indicates the probability of that region to belong to a specific class. For the region generation process, a small network iterates the image with an input of an $n \times n$ region, using a *sliding-window* approach. This window is then taken as input by a convolutional network with a $n \times n$ size followed by two more layers, the box-regression layer, and the box-classification layer. Centered around each window there are anchors that are related with the scale and aspect ratio, the formula that normally defines the number of anchors on each image is $Width \times Height \times 9$. Anchors are used as a reference for the region features without having to contain the region itself, making it easier to pass it to the Fast R-CNN lateron. The number of regions proposals is reduced, applying the non-Maximum suppression method based on the scores of each region. This method reduces the number of regions that overlap other ones which have been scored a highervalue, that way the same area is kept covered but at the same time, the numberof proposals is reduced.

### 3.7 Model application and assessment

When applying the model, one value is important to keep in mind, the threshold (MODEL.ROI_HEADS.SCORE_THRESH_TEST). When executed, *Detectron2* may give each detection a numeric value in the form of *confidence* parameter, which somehow gives an estimation for the detected defect to be a true positive (TP). If we set this value too low, the results would present a high number of non-secure detections that would not be useful at all. Finally, we used a value of 0.7 for this parameter.

Results will show our original image with a new layer over it containing the detection results. In particular:

– **Segmentation area:** Defect area predicted by the model.
– **Bounding box:** Square-shape line that will cover the segmentation area.
– **Detection name:** *Detectron2* will generate a name from the metadata of the COCO file, and we therefore could easily differentiate different types of defects, if they were predicted.
– **Confidence percentage:** A *probability* of being a true positive for the model.

During the training stage, when 600 iterations were reached, there was no detection improvement and the validation loss started to increase, meaning that the overfitting phenomenon started to take place. It was at this point that the model started to overlap some detections and increase the number of false positives (FP). Training then stopped.

The detection output consists of several numeric values contained in different structures. All these values can be easily extracted and converted to simple numeric values and then, turned into ground truth image of the original image exactly as the masks used for the generation of the COCO file and annotation of the initial images. This allows to compare the original binary mask with the new

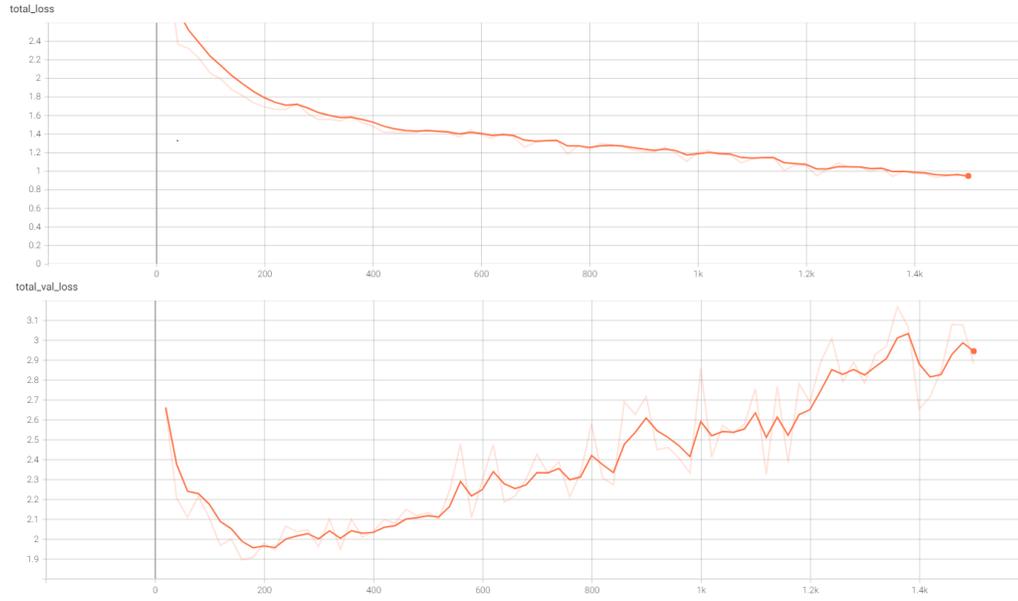

**Fig. 3.** Training and validation loss evolution

generated one. Three criteria were used to assess the method: Precision, Recall and F1.

- **Precision**: It allows to quantify how many True Positives present on the results are intended to be so according to the original Ground Truth images. Its formula is:

$$\text{Precision} = \frac{TP}{TP + FP}. \qquad (1)$$

- **Recall**: It represents how many values, Positive or Negative, are covered in the results and are not *ignored* by the model. Its formula is:

$$\text{Recall} = \frac{TP}{TP + FN}. \qquad (2)$$

- **F1**: Combines precision and recall, and makes it easier to compare results with other algorithms. Usually known as harmonic mean between the previous two. The F1 formula is:

$$\text{F1} = \frac{2 \cdot (\text{Recall} \cdot \text{Precision})}{(\text{Recall} + \text{Precision})}. \qquad (3)$$

Figure 4 shows how the increase in the number of iterations affected the model behavior. This number was increased by 100 more than the previous step, and executed over the test dataset, obtaining a graphic evolution of the three metrics mentioned in the previous section. We could see that around 600 to 700 iterations marked the point where the improvement in the metrics values seemed to stop, and created a subsequent overfitting phase. This resulted in an increasing number of False Positives detected on each image.

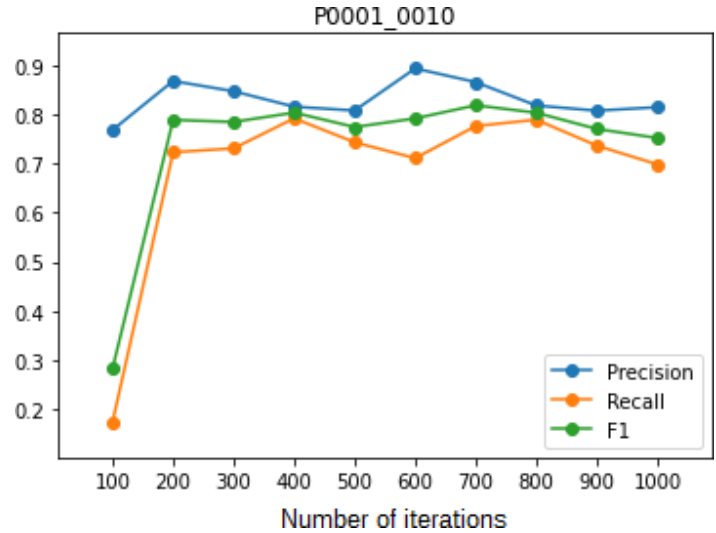

**Fig. 4.** Evolution of the three metrics with with the number of iterations

### 3.8   Results and discussion

Figure 5 shows the defect detection results obtained by the Faster R-CNN method. In particular, Figure 5 (left column) shows the X-Ray original im- ages; Figure 5 (right column) shows the defect detection results obtained by theFaster R-CNN method. We can see that the method correctly detects most of the defects, and we can also see how irregular (in shape), complex and difficultto detect these defects might be.

Figure 6 shows false positive detections, which present how the model fails to discard some elements that present an important similarity both in shape and grayscale distribution to the defect it has been trained with, marking them as defects as well. However, when running this kind of test some true positive defects that we did not show at first and for that reason tested it as a false positive image, were indeed detected by the model. Specially the smaller size ones.

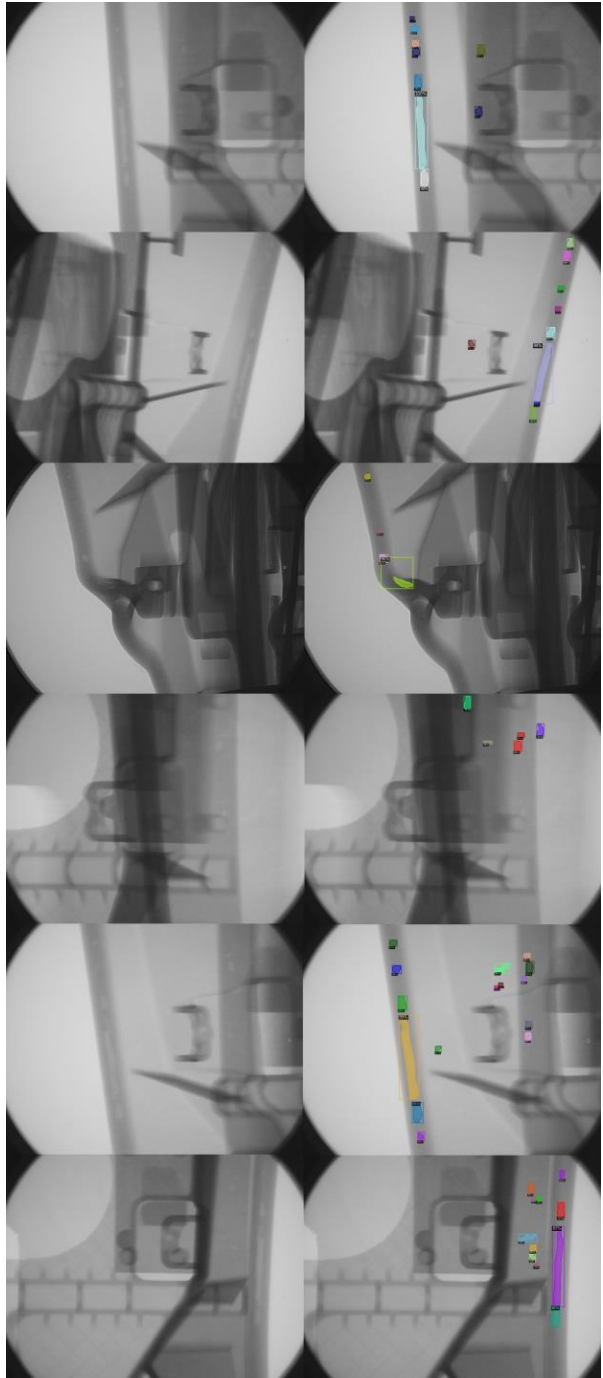

**Fig. 5.** (Left) X-Ray original images; (Right) defect detection results obtained by the Faster R-CNN method.

## 4    Conclusions

In this paper, we have proposed the application of a deep learning methodology based on *Detectron2* for the identification and segmentation of these defects on X-ray images obtained mainly from automotive parts. Qualitative results show that this tool is able to identify almost any type of defect (whether it is small or big, and of shape changing nature) in them. It also delineates their (irregular and complex) boundaries. Taking into account their varied shape and structure, they could be categorized into small, medium-sized or big sized defects. Future work will include the possibility to obtain images of defects from different perspectives, in order to create a 3D structure of the pieces (and the defects) and be then able to apply their quality criteria based on depth and extension of the defect, and therefore on its severity.

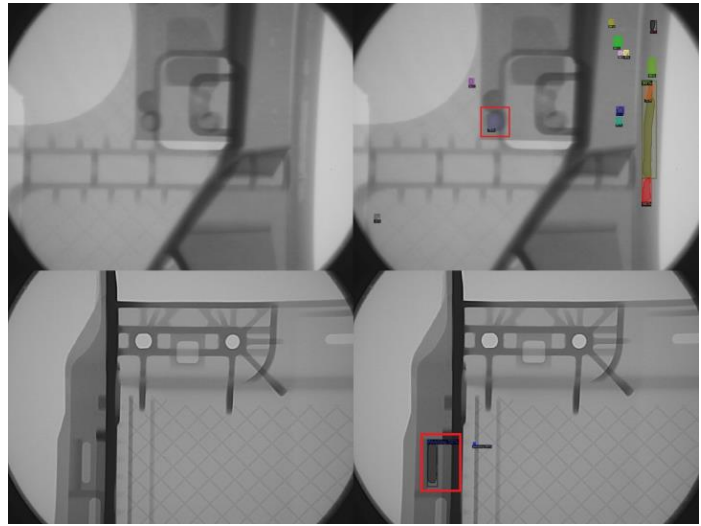

**Fig. 6.** False-Positives cases that took place occasionally or as an overfitting symptom.